# Fast energy-aware OLSR routing in VANETs by means of a parallel evolutionary algorithm

Jamal Toutouh · Sergio Nesmachnow ·
Enrique Alba

**Abstract** This work tackles the problem of reducing the power consumption of the OLSR routing protocol in vehicular networks. Nowadays, energy-aware and green communication protocols are important research topics, specially when deploying wireless mobile networks. This article introduces a fast automatic methodology to search for energy-efficient OLSR configurations by using a parallel evolutionary algorithm. The experimental analysis demonstrates that significant improvements over the standard configuration can be attained in terms of power consumption, with no noteworthy loss in the QoS.

## 1 Introduction

In the last five years, the networking research community has shown a growing interest in vehicular ad hoc networks (VANETs), a technology that uses vehicles as nodes of a mobile network [24]. VANETs share their main concepts with generic mobile ad hoc networks (MANETs), but they also have several distinctive features. For example, the node mobility in VANETs is different from the models used in other mobile networks, since vehicles tend to move following organized patterns, and they are usually subject to restrictions in both their motion range and in the interactions with roadside infrastructure. In addition, VANETs integrate multiple ad hoc networking technologies (such as WiFi IEEE 802.11p, WiMAX IEEE 802.16, Bluetooth, etc.), posing a difficult challenge for attaining effective and simple communication between vehicles.

S. Nesmachnow
Universidad de la República, Uruguay
E-mail: sergion@fing.edu.uy

J. Toutouh, E. Alba
University of Málaga, Spain
E-mail: {jamal,eat}@lcc.uma.es

VANETs involve communication between vehicles and other battery-fed devices—pedestrian smartphones, road transceivers, sensors—. Thus, the power consumption by wireless communications becomes a major concern, and the use of energy-efficient communications is highly desirable.

Network routing is a critical issue in VANETs, as well as in any other ad hoc network. The absence of a central entity to manage the routing information, the limitations of the shared medium, and the dynamic topology due to the high node mobility and obstacles, make the routing problem even harder. Proactive protocols are a useful choice for routing in VANETs, since they generally outperform reactive ones in terms of quality of service (QoS), network throughput, and end-to-end delay [23]. However, proactive protocols have a higher routing overhead, significantly reducing their energy efficiency [8, 31].

Optimized Link State Routing (OLSR) [12] is a well-known proactive routing protocol used in VANETs. The energy efficiency of OLSR has been studied focusing on specific protocol variants [13, 27], but VANET infrastructures have seldom been considered. The OLSR power consumption can be improved by modifying the standard parameter configuration, in order to reduce the routing overhead. However, is not easy to find the best OLSR configuration. Exact and enumerative methods are not applicable to solve the underlying optimization problem, since they require prohibitive execution times to perform the search, even when considering only a small set of parameter values. In this context, metaheuristics are a promising option to find accurate energy-aware OLSR configurations in reasonable times, even when a large set of parameter values are considered, as in the problem tackled in this paper.

Evolutionary algorithms (EAs) have emerged as flexible and robust metaheuristics for search and optimization, achieving a high level of problem solving efficacy in many application areas [6]. In order to further improve the efficiency of EAs, parallel implementations have been used to significantly enhance and speed up the search, allowing high quality results to be computed in reasonable execution times even for hard-to-solve optimization problems [1].

This work proposes applying an automatic configuration of the main OLSR parameters by using a parallel EA. The main goals of the research are: i) to improve the efficiency of OLSR in VANETs, trying to reduce the power consumption when using the standard Request for Comments (RFC) 3626 configuration [12], and ii) to scale down the times required to perform the automatic configuration, in order to study large realistic VANET scenarios.

The methodology applied in this work consists of exploring the search space for possible combinations of eight parameter values that define the OLSR routing protocol, by using a genetic algorithm (GA). The power consumption due to data exchange of each OLSR configuration is evaluated using the data obtained after performing VANETs simulations with the ns-2 network simulator. Since these simulations require a long time to perform, a parallel implementation of the GA is used in order to reduce the search execution times. The best configurations are compared with the standard one defined by RFC 3626, both in terms of power consumption and QoS. Finally, the best energy-aware OLSR configuration found is validated on a set of 36 VANET scenarios.

The article is organized as follows. Section 2 introduces the energy-aware routing problem in VANETs, the OLSR protocol, the power consumption model, and reviews related work on metaheuristics for protocol optimization in MANETs/VANETs and methods for energy-efficient OLSR. Section 3 describes evolutionary computing and the parallel model for EAs employed here. Section 4 presents the implementation details of the parallel GA to find energy-aware OLSR configurations in VANETs. The experimental analysis in Section 5 studies the numerical efficacy and the computational efficiency of the parallel GA, and also presents a validation of the best configuration found on a large set of VANET scenarios. Finally, Section 6 presents the main conclusions of the research and formulates the main lines for future work.

## 2 Energy aware routing in vehicular networks

This section introduces VANET routing, the OLSR protocol, the power consumption model used in our approach, and a review of related work. It also describes the methodology for finding energy-efficient OLSR configurations.

### 2.1 Routing in VANETs

Finding a stable routing strategy that guarantee the exchange of up-to-date information, maximizing reliability and minimizing delays is an important technical challenge when designing an architecture for vehicular communication.

In VANETs, the links for vehicle-to-vehicle and vehicle-to-infrastructure communication tend to be shortlived, due to the intrinsic high-speed node mobility and the presence of obstacles. Therefore, a great deal of effort is dedicated to defining efficient routing strategies. Specific VANET protocols have appeared over the last few years, but most of them are based on prior mobile ad hoc networks. These protocols can be grouped into: *topology-based* (proactive, e.g. DSDV and OLSR, reactive, e.g., AODV and DSR, hybrid), *position-based* (e.g., GPSR, GEOTORA, GPCR), *cluster-based* (e.g., COIN, LORA CBF) and *broadcasting* (e.g., BROADCOMM, V-TRADE, HV-TRADE) [30, 29].

Within those protocols originally proposed for MANETs, topology-based protocols are among the most studied for routing in VANETs [29]. In proactive protocols, all nodes have consistent and up-to-date routing information for each node permanently, unlike in reactive ones, where the routes are created when demanded by the source node [30]. Proactive protocols have the advantage of reduced end-to-end delays, since the routes are already established and it is not necessary to invoke a routing discovery process to find them, as in reactive protocols. However, proactive protocols require a continuous exchange of control messages to maintain the topological information stored in the routing tables. While negligible for small scenarios, control messages use significant additional bandwidth for large networks, leading to excessive power consumption, possibly preventing the use of devices fed by batteries or renewable energy sources in VANETs [23].

In this work, we restrict our attention to OLSR, a proactive routing protocol that has been analyzed for use in VANETs through both simulations [9, 28] and real world tests [38]. In turn, different comparisons of this protocol against a reactive approach (AODV) concluded that OLSR principally outperforms AODV in terms of delivery delays and path lengths, while keeping a similar percentage of packets delivered correctly [23, 39]. The type of routing protocol affects the nodes power consumption in two different ways: the routing network load influences the amount of energy used to send and receive routing control messages; and the generated routing paths affect the power consumption in those nodes forwarding the packets [8, 40].

For the aforementioned reasons, we have selected OLSR as case-of-use, as its main drawback is the power consumption. Thus, we can analyze the use of our parallel GA to deal with the energy-efficiency routing problem in VANETs.

## 2.2 Optimized link state routing protocol

OLSR is a proactive link-state routing protocol conceived for mobile ad hoc networks with low bandwidth and high mobility. OLSR relies on applying an efficient periodic flooding of control information using special nodes that act as *multipoint relays* (MPRs), reducing the number of required transmissions [32].

OLSR daemons periodically exchange control messages to maintain the network topology information in the presence of mobility and failures. The core functionality is performed mainly by using three different types of messages:

- HELLO messages, exchanged between neighbor nodes to allow for link sensing, neighborhood detection, and MPR selection signaling. These messages are generated periodically, containing information about the neighbor nodes and about the links between their network interfaces.
- TC (topology control) messages, generated by MPRs to indicate which other nodes have selected it as their MPR. This information is used for routing table calculations. TC messages are broadcasted periodically, and a sequence number is used to distinguish between recent and old ones.
- MID (multiple interface declaration) messages, sent by the nodes to report information about their network interfaces, needed since multiple interfaces with different addresses can be involved in the communications.

OLSR is regulated by a set of parameters defined in the OLSR RFC 3626 [12]:

- the timeouts before resending each message type, HELLO INTERVAL, REFRESH INTERVAL, and TC INTERVAL, respectively;
- the "validity time" of the information received for each message type, NEIGHB_HOLD_TIME, MID_HOLD_TIME, and TOP_HOLD_TIME;
- the WILLINGNESS of a node to act as a MPR;
- the time during which the MPRs record information about the forwarded packets, DUP_HOLD_TIME.

A set of default values for these parameters has been suggested by the OLSR standard RFC 3626 (see Table 1).

Table 1  Main OLSR parameters and standard values in the RFC 3626.

| parameter | standard value (RFC 3626 [12]) | range |
|---|---|---|
| HELLO_INTERVAL | 2.0 s | $R \in [2.0, 15.0]$ |
| REFRESH_INTERVAL | | $2.0 sR \in [2.0, 15.0]$ |
| TC_INTERVAL | 5.0 s | $R \in [4.0, 35.0]$ |
| WILLINGNESS | 3 | $Z \in [0, 7]$ |
| NEIGHB_HOLD_TIME | $3 \times$ HELLO_INTERVAL | $R \in [5.5, 45.0]$ |
| TOP_HOLD_TIME | $3 \times$ TC_INTERVAL | $R \in [10.5, 90.0]$ |
| MID_HOLD_TIME | $3 \times$ TC_INTERVAL | $R \in [10.5, 90.0]$ |
| DUP_HOLD_TIME | 30.0 s | $R \in [10.5, 90.0]$ |

OLSR has several features that make it suitable for highly dynamic ad hoc networks as VANETs: i) it is well suited for high density networks, with concentrated communication between a large number of nodes [12, 25]; ii) it is useful for applications requiring short delays in the data transmission, as most of warning information in VANETs [25]; iii) the protocol information can be extended with data to allow the hosts to know in advance the quality of the routes; iv) it permits an easy integration into existing operating systems and devices, including smartphones, embedded systems, without changing the header of the IP messages [19]; and v) it manages multiple interface addresses for the same host, allowing VANET nodes to use different network interfaces— WiFi, Bluetooth, etc.—, while acting as gateways to other devices, such as drivers and pedestrian smartphones, base stations, etc. [12].

## 2.3 Power consumption model

Several agents are involved in VANET communications, such as on-board devices, smartphones, or traffic signs, which use wireless network interfaces to exchange information with each other. The energy required for each device to perform the communications depends on its mode:

— *idle* is the default state of wireless interfaces in ad hoc networks, where nodes keep listening and the interface can change the state and start transmitting or receiving packets.
— *transmit* and *receive* states are for sending and receiving data through the medium.
— *sleep* state is when the node radio is turned off, and thus the node is not capable of detecting any signal.

In our work, we modify the behavior of OLSR in order to reduce the power consumption due to data exchange (control or information messages). We deal with energy-awareness in VANETs by optimizing the power consumption of the two operational states that act during the packet exchange: transmit and receive states. Therefore, we consider the *per-packet power consumption* [16] modeled by Cano et al. [8], in which only transmit and receive modes are taken into account to compute the power consumption to be optimized.

The energy is computed according to the power requirements in transmitting ($P_{send}$) and receiving ($P_{recv}$) states, and the time needed to transmit the packets (*time*). These values are obtained by using the network interface card (NIC) characteristics of electric current ($I_{send}$, $I_{recv}$) and power supply ($V_{send}$, $V_{recv}$) in each state, the size of the packets, and the bandwidth used.

Equations 1 and 2 represent the energy required for packet transmission ($E_{send}$) and for packet reception ($E_{recv}$).

$$E_{send} = P_{send} \times time = (I_{send} \times V_{send}) \times \frac{PacketSize}{Bandwidth} \qquad (1)$$

$$E_{recv} = P_{recv} \times time = (I_{recv} \times V_{recv}) \times \frac{PacketSize}{Bandwidth} \qquad (2)$$

According to the specification of the Unex DCMA-86P2 NIC [43] modeled in our simulations, the power consumption is from 440 mA in transmitting mode, and from 260 mA in receiving mode, and it is fed with 5.0 V. This network interface uses a 6 Mbps bandwidth implementation of the standard IEEE 802.11p. Thus, the power consumption in transmitting ($E_{send}$) and receiving states ($E_{recv}$), in Joules, are given by Equations 3 and 4, respectively, where the packet size is given in bits.

$$E_{send} = (440 \times 5) \times \frac{PacketSize}{6 \times 10^6} \qquad (3)$$

$$E_{recv} = (260 \times 5) \times \frac{PacketSize}{6 \times 10^6} \qquad (4)$$

The total power consumption for a packet transmission is the sum of the costs incurred by the sending node and all receivers, whether they are the destination nodes or not. Equation 5 computes the total power consumption per packet ($E_{total}$) when there are $r$ receiver nodes in the communication range of the sender.

$$E_{total} = E_{send} + \sum_{i=1} E_{recv} \qquad (5)$$

## 2.4 Related work

The need of providing efficient communications in MANETs and VANETs has motivated the research community to deal with the problem of optimizing the communication protocols employed in such networks. The related studies have mainly focused on obtaining dramatic improvements in terms of both QoS offered—packet delivery ratio, delivery delays, etc.—and resources consumed, e.g. power requirements. Due to the complexity of the underlying optimization problems, metaheuristics have been usually applied as the most appropriate techniques to solve them.

### 2.4.1 Metaheuristics for protocol optimization in MANETs and VANETs

Regarding optimization techniques in MANETs, Alba et al. [3] applied a specialized cellular multi-objective GA for finding an optimal configuration for the Delayed Flooding with Cumulative Neighborhood broadcasting strategy. Dorronsoro et. al [15] evaluated six different versions of GA for the design of ad hoc injection networks. Cheng et al. [10] also used a GA for dealing with  the multicast routing problem in MANETs. More recently, Ruiz et al. [35, 36] applied a hybrid multi-objective optimization algorithm (CellDE) to maximize the coverage and minimize the power consumption and broadcast time of the EDB protocol.

In VANETs there are just a few approaches applying metaheuristics to optimize communication protocols. Garcı́a-Nieto et al. [18] employed a set of metaheuristic algorithms to optimize VDTP and AODV [17] protocols. Recently, Toutouh et al. [42] applied DE in order to improve the performance of OLSR routing protocol in such networks.

### 2.4.2 Methods for energy-efficient OLSR

The related literature presents a number of power-aware mechanisms proposed at the network layer in wireless networks, mainly due to the impact of the routing protocols on the overall power consumption. These protocols determine the power consumption in creating and maintaining the routes and the data packets forwarding. In this work, we aim to provide an energy-efficient OLSR configuration when applying this protocol for routing in VANETs. OLSR has been selected as a case study since it offers competitive QoS in such networks [41], but it also requires significant power consumption.

Several approaches have been proposed to reduce the power consumption when using OLSR. Ghanem et al. [20] and Razalli et al. [33] evaluated new MPRs selection criteria based on the residual energy levels of the nodes. Routing path determination based on the overall power consumption to forward data and on the residual level of energy of intermediate nodes was explored  by De Rango and Fotino [14] and Guo and Malakooti [22], respectively. Other authors have analyzed combinations of the aforementioned techniques [7, 13, 27, 31, 37]. Finally, De Rango et al. [13] presented *Overhearing Exclusion*, a mechanism that allows energy saving by turning off the device when a unicast message exchange happens in the device neighborhood.

Our previous article [40] studied the possible energy savings when an efficient protocol configuration in terms of QoS (DE-OLSR) is used. That is the only existing work studying the best parameter configurations to improve the energy efficiency of OLSR specifically in VANETs. The impact of the parameters configuration in the network performance led us to perform the in-depth study of the OLSR parameter tunning that we now present, in order to find the best configuration in terms of energy efficiency in VANETs. As in previously presented MANET/VANET optimization problems, the use of metaheuristic techniques is mandatory to deal with such problems.

## 2.5 Methodology for energy efficient OLSR via parameter tuning

The standard OLSR parameter values in Table 1 can be fine-tuned automatically by using an optimization technique, with the aim of obtaining efficient OLSR configurations for VANETs. This procedure hopefully allows reducing the power consumption without incurring a significant loss of QoS in comparison with the standard OLSR definition in RFC 3626.

The search of possible combinations of OLSR parameter values is not an easy problem. The dimension of the search space increases exponentially with both the number and the range of possible parameter values. Thus, exact search methods are not useful for efficiently solving the problem. In this context, heuristic and metaheuristic optimization algorithms are viable options to compute accurate energy-aware configurations in reasonable times.

In our previous paper [42], the large amount of time required to perform the VANET simulations limited the proposed search method to work with a reduced population in order to obtain results in reasonable time. To overcome this drawback, this work proposes to use a parallel GA for efficiently searching the parameter values of the OLSR protocol. By using several computing resources simultaneously, the parallel implementation allows to reduce the simulation times.

The automatic search for energy-aware OLSR configurations is carried out by using the energy cost of the communications as the main objective to be optimized. However, since excessive reductions of power consumption of the protocol can cause it to malfunction, we use the *packet delivery ratio* (PDR) quality metric to guarantee a minimum level of QoS in the communications. Thus, the parallel GA for finding energy-efficient parameter values searches the *best* configuration that provides the most energy savings while maintaining PDR within margins of good performance (the degradation in the PDR value is kept below 15% of the PDR achieved with the standard OLSR configuration).

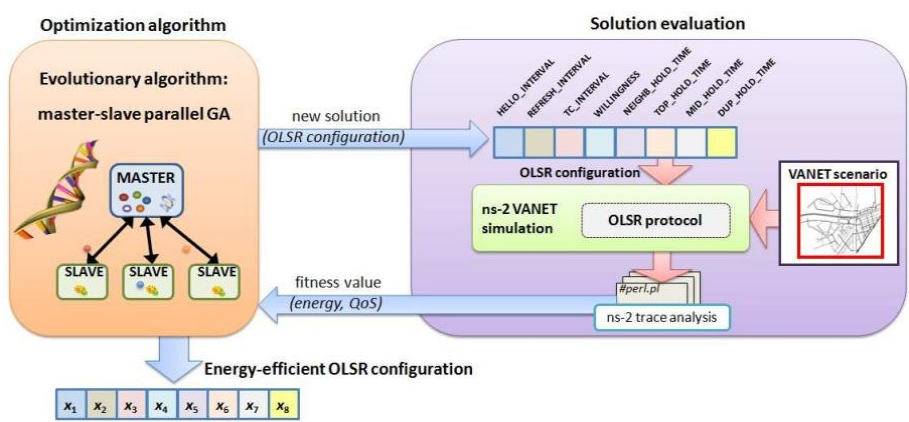

Fig. 1 Automatic methodology for energy-aware OLSR tunning.

Fig. 1 summarizes the automatic methodology for finding energy-aware parametrizations for the OLSR protocol in VANETs. The proposed method integrates the evolutionary search via a parallel GA, the routing simulation in VANETs using the ns-2 network simulator and the UM-OLSR implementation from University of Murcia [34], and a set of scripts developed to evaluate the power consumption and QoS using the ns-2 output.

## 3 Evolutionary computation

This section introduces the main concepts about evolutionary computation and the parallel model applied to the GA used in this paper.

### 3.1 Evolutionary algorithms

EAs are non-deterministic methods that emulate the evolutionary process of species in nature, in order to solve optimization, search, and other problems [6]. Over the last twenty years, EAs have been successfully applied for solving optimization and search problems underlying many complex real-life applications. An EA is an iterative technique (each iteration is called a *generation*) that applies stochastic operators on a pool of individuals (the population). Each individual in the population is the encoded version of a tentative solution of the problem. The initial population is generated either by using a random method or by applying a specific heuristic for the problem.

An evaluation function associates a *fitness* value with every individual, indicating its suitability to the problem. Iteratively, the population is modified by the probabilistic application of *variation operators* like the *recombination* of individuals or random changes (*mutations*) in their contents. A selection technique that gives a higher chance of survival to the best suited individuals, guides the EA to tentative solutions of better quality through the generations.

The stopping criterion usually involves a fixed number of generations or execution time, a quality threshold on the best fitness value, or the detection of a stagnation situation. Specific policies are used to select the individuals to recombine and to determine which new individuals are inserted in the population in each new generation. The EA returns the best solution ever found in the iterative process, taking into account the fitness function considered.

The classic GA [21] is an EA that defines recombination and mutation as variation operators, applying them to the population of potential solutions in each generation. The recombination is used as the main operator to perform the search (exploiting the characteristics of suitable individuals), while the mutation is used as a (seldom-applied) secondary operator aimed at providing diversity for exploring different zones of the search space.

GAs are widely spread due to their versatility for solving optimization problems. Here, a parallel version of the classic GA has been applied to the problem of finding energy-aware OLSR configurations in VANETs.

## 3.2 Parallel evolutionary algorithms

Parallel implementations became popular in the last decade as an effort to improve the efficiency of EAs. By splitting the population or the fitness function evaluation into several processing elements, parallel EAs allow reaching high quality results in a reasonable execution time even for hard-to-solve optimization problems [2]. The parallel GA proposed here is categorized within the *master-slave* model according the classification by Alba and Tomassini [4]. The master-slave model (see Fig. 2) follows a classic functional decomposition of the EA, where different stages of the evolutionary process are performed in several computing resources. The evaluation of the fitness function is the main candidate to perform in parallel, since it usually requires larger computing time than the application of the variation operators.

Thus, a master-slave parallel EA is organized in a hierarchic structure: a master process performs the evolutionary search and controls a group of slave processes that evaluate the fitness function.

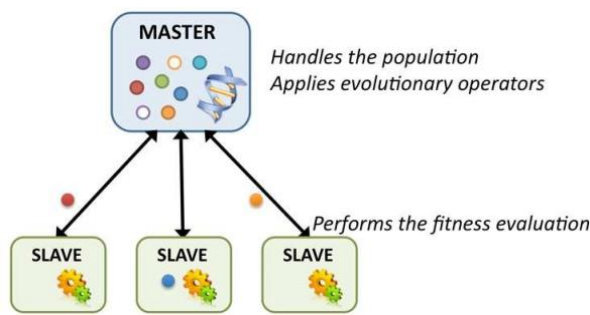

Fig. 2 Master-slave model for parallel EAs.

## 4 A parallel GA for energy-aware OLSR tunning

This section presents the implementation details of the parallel GA designed to find the energy-aware configuration of OLSR.

### 4.1 The MALLBA library

MALLBA [2] is a library of optimization algorithms that deals with parallelism in a user-friendly and efficient manner. MALLBA implements EAs and other metaheuristics as generic templates in *software skeletons* to be instantiated with the problem features by the user. These templates incorporate the knowledge related to the resolution method, its interactions with the problem, and the parallelism. Skeletons are implemented by *required* and *provided* C++ classes that abstract the entities in the resolution method:

- The *provided classes* implement internal aspects of the skeleton in a problemindependent way. The most important provided classes are Solver that implements each optimization algorithm, SetUpParams for setting the algorithms' parameters, and Population to store a set of candidate solutions.
- The *required classes* specify information related to the problem. Each skeleton includes the Problem and Solution required classes, that encapsulate the problem-dependent entities needed by the resolution method.

## 4.2 Parallel multithreading GA in MALLBA

The skeletons in MALLBA offer support for parallelism using the distributed memory approach (i.e., implementing distributed subpopulation models for metaheuristics). However, the library does not provide support for sharedmemory multithreading parallel programming.

Multithreading programming allows implementing efficient algorithms by using multiple threads within a single process. Multihtreading is well suited for multi-core computers, where each thread is executed on a single core. It provides a fast method for concurrent execution; communications and synchronizations are performed via the shared-memory resource, which is handled using mutually-exclusive operations in order to prevent simultaneous accesses. There is a runtime overhead for creating and destroying threads, and a common approach to avoid it is using a *thread pool*. Instead of creating a new thread, the application uses an available thread from the pool, performs its task, and returns the thread to the pool instead of destroying it. This reusing methodology improves the performance of the parallel program, by reducing the cost of performing the creation and termination of threads.

The multithreading master-slave parallel EA proposed in this work was implemented using the GA skeleton in MALLBA. Additional code was incorporated into the GA skeleton to implement several new features:

- to create and manage the pool of threads used for the fitness evaluation;
- to implement the master-slave hierarchy and the communications between master and slaves;
- to define the synchronization mechanisms between threads, used to read and write the shared memory.

Our implementation starts by creating and initializing a pool of threads to distribute the fitness evaluation. Each thread receives several input parameters from the master process, including the solution to be evaluated, the thread identification, and the index in the array of fitness values. Then, each slave process, implemented in each thread, computes the fitness evaluation by simulating the mobile communications with the proposed OLSR parameters configuration in a given VANET scenario, using the ns-2 network simulator. The master process, implemented in the main thread of execution, is in charge of performing the domain decomposition for the problem, by assigning each thread the solutions to be evaluated. After that, the master process waits until all slave threads finish their execution and report the fitness value.

### 4.3 Problem encoding

The OLSR protocol is governed by eight different configuration parameters, already presented in Table 1. For this reason, in the parallel GA the solutions are represented by individuals encoded as a vector with eight genes, one for each parameter, as presented in Fig. 3.

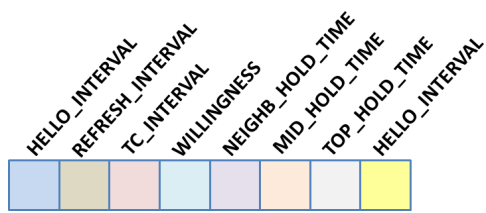

Fig. 3 Solution encoding for the energy-aware OLSR tunning problem.

The first three genes are real valued, and they represent the timeout timers before resending control messages (HELLO INTERVAL, REFRESH INTERVAL, and TC INTERVAL, respectively). The forth one encodes the WILLINGNESS parameter, and therefore, it takes an integer value from zero to seven. Finally, the last four genes are real valued, and they denote the timeout hold timers of OLSR (NEIGHB HOLD TIME, MID HOLD TIME, TOP HOLD TIME, and DUP HOLD TIME, respectively). The valid ranges for each one of the gene values have already been presented in Table 1.

### 4.4 Fitness function

The fitness function is crucial for the GA optimization mechanism, since it guides the population to solutions of better quality. The optimization proposed in this work mainly concerns to energy-aware communications, so the main component of the fitness function is the power consumed by the VANET nodes when using a certain OLSR configuration. However, if a given configuration excessively reduces the power consumption, the protocol may not satisfy the QoS requirements for the communication in VANET networks. So, there is a tradeoff between the energy efficiency and the QoS provided by the protocol. In order to take into account the previous consideration, the fitness function used in the parallel GA proposed in this work integrates the PDR metric in order to bias the search to solutions with acceptable QoS.

The fitness function is given by the expression in Equation 6, where $E(s)$ and $PDR(s)$ represent the power consumption and the PDR for a given OLSR configuration $s$, respectively. $E_{RF\ C}$ and $PDR_{RF\ C}$ are the reference values for the power consumption and the PDR when using the standard configuration in RFC 3626, respectively. Finally, $\omega_1 = 0.9$ and $\omega_2 = -0.1$ are the weights for the energy and PDR contributions, respectively, and $\Delta=0.1$ is a normalizing offset to keep the fitness value in the interval [0, 1].

$$F(s) = \Delta + \left( \omega_1 \times \frac{E(s)}{E_{RFC}} + \omega_2 \times \frac{PDR(s)}{PDR_{MAX}} \right) \qquad (6)$$

Equation 6 is valid for solutions with a PDR degradation lower than 15% of the reference PDR value. In order to keep in the GA population those solutions with still a lower PDR, but containing potentially useful genetic information, the penalization model in Equation 7 was applied.

$$F_P(s) = F(s) + \left( (0.85 \times PDR_{RFC} - PDR(s)) \times \frac{E(s)}{E_{RFC}} \right) \qquad (7)$$

The penalized fitness $F_P(s)$ takes into account the gap between the PDR of the evaluated solution and the worst PDR value admitted ($0.85 \times PDR_{RFC}$), and the ratio between the energy of the evaluated solution and the reference energy value $E_{RFC}$.

### 4.5 Parallel GA operators

A classic GA has been applied for protocol tuning in a previous paper [18]. However, although it offered competitive results, that algorithm suffered from low population diversity and early stagnation. For this reason, in this work we decided to introduce some variations to the canonical *initialization* and *mutation* operators.

#### 4.5.1 *Initialization*

The population initialization should distribute the individuals uniformly in the search space as much as possible. However, this uniform pattern is not easy to obtain when using random operators and small populations. Therefore, here we propose using a uniform initialization, to ensure that the initial population contains individuals from different areas of the parameters' search space. The initialization operator splits the search space into *pop_size* diagonal subspaces (where *pop_size* is the global population size of the parallel GA), and it forcibly ensures that there is an individual located in each diagonal subspace [11]. Equation 8 summarizes the procedure applied in the initialization operator.

$$x_{p,i}^{(0)} = z_i^{RFC} + \alpha^p \quad i \in [0, 7], p \in [0, pop\_size - 1] \qquad (8)$$

In Equation 8, $x_{p,i}^{(0)}$ is the initial value for each gene $i$ in the solution vector that encodes the $p$-th individual, set according to a *population seed* $z_i^{RFC}$ and a randomly distributed value $\alpha^p$. $z_i^{RFC}$ is the value proposed by the RFC 3626 for the $i$-th OLSR parameter. $\alpha^p$ is computed by using the diagonal subspace limits and a random value $\beta \in [0, 1]$, as expressed in Equation 9, where $z_{(i,MAX)}$ and $z_{(i,MIN)}$ are the upper and lower values for the $i$-th parameter, according to the ranges defined in Table 1.

$$\alpha^p = \left( \frac{p + \beta}{pop\_size} \right) \times (z_{(i,MAX)} - z_{(i,MIN)}) \qquad (9)$$

### 4.5.2 *Recombination*

The parallel GA uses the classic arithmetic recombination operator for realvalued problem encodings. It defines a linear combination of two chromosomes, $x_p^{(g)}$ and $x_q^{(g)}$, according to Equation 10, where the best parent governs the reproduction according to the weight $\sigma \in [0, 1]$.

$$x_{p,i}^{(g+1)} = \sigma \times x_{p,i}^{(g)} + (1 - \sigma) \times x_{q,i}^{(g)}$$
$$x_{q,i}^{(g+1)} = (1 - \sigma) \times x_{p,i}^{(g)} + \sigma \times x_{q,i}^{(g)} \qquad (10)$$

### 4.5.3 *Mutation*

The mutation operator introduces new genetic information, and therefore, diversity to the population of the parallel GA. After analyzing the algorithm of the OLSR protocol, we decided to introduce some problem-related information in the mutation operator. Thereby, the new genetic information is randomly generated, but it does not represent pointless OLSR configurations. The genes that encode OLSR related parameters, e.g., HELLO INTERVAL and NEIGHB HOLD TIME [12], are modified simultaneously, but using different policies and following the OLSR power-aware problem specifications. According to this idea, the mutation operator offers 22 different movements in the solution space. For example, Equation 11 presents the case in which HELLO INTERVAL ($x_{p,0}^{(g)}$) and NEIGHB HOLD TIME ($x_{p,4}^{(g)}$) genes are mutated in generation $g$. A similar procedure is employed for other parameters.

$$x_{p,0}^{(g+1)} = \beta_0 \times (z_{(0,MAX)} - z_{(0,MIN)}) \quad \beta_0 \in [0, 1]$$
$$x_{p,4}^{(g+1)} = \beta_4 \times (z_{(4,MAX)} - z_{(4,MIN)}) \quad \beta_4 \in [0, 1] \qquad (11)$$

## 5 Experimental analysis

This section introduces the set of VANET scenarios and the computational platform used to evaluate the proposed parallel GA. After that, the experiments to determine the best values for the GA parameters are presented. First, the experimental results when solving realistic VANET scenarios are analyzed, by presenting the numerical results and a comparative analysis of solution quality and computational efficiency when using a different number of threads. Last, the best solutions found are validated by studying their performance on a set of 36 VANET scenarios.

## 5.1 VANET scenarios

The experimental evaluation of the proposed parallel GA was performed using urban VANET scenarios covering real areas of the city of Málaga, Spain.

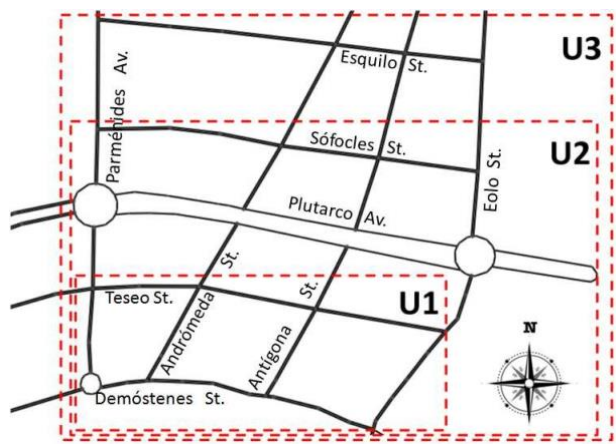

Fig. 4 M´alaga urban areas taken into account in each VANET scenario.

A total number of 36 scenarios were used, considering the three areas shown on the map in Fig. 4.

In the first stage, the simulations in the parallel GA parameter setting experiments were done in a small-sized scenario (U1) with 20 vehicles moving along the roads. The optimization of OLSR parameters using parallel GAs was performed using a medium-sized scenario (U2), also with 20 vehicles. Lastly, in the validation experiments, 36 scenarios with different area sizes, traffic densities (number of vehicles per square meter), and communication patterns were used. In each case, realistic simulation mobility models were generated using the open source traffic simulation package SUMO [26], where vehicles move following real traffic rules (traffic lights and signs) during 180 s.

The VANETs were evaluated by using the ns-2 network simulator, having nodes configured following the 'Standard Wireless Access in Vehicular Environments" (WAVE). In order to evaluate the performance of the routing protocol, different constant bit rate (CBR) traffic sources were randomly chosen to generate the packets that travel through the network.

Table 2 presents the main features of the VANET scenarios and the network specification used in the experimental analysis. All the scenarios, mobility models and network workloads used are publicly available to download in http://neo.lcc.uma.es/vanet/download-simulations.

## 5.2 Development and execution platform

The parallel GA was implemented in C++, using MALLBA and the standard pthread library. The experimental analysis was performed in a cluster with Opteron 6172 Six-Core processors at 2.1 GHz, with 24 GB RAM, CentOS Linux, and Gigabit Ethernet (Cluster FING, Facultad de Ingenier´ia, Universidad de la República, Uruguay; cluster website: http://www.fing.edu.uy/cluster).

Table 2  Details of the VANET scenarios and network specification.

| scenario | area size | vehicles | CBR sources | | parameter | value/protocol |
|----------|-----------|----------|-------------|---|-----------|----------------|
| U1 | | | | | Propagation model | Nakagami fading |
| | | | | | Max. radio range | 500 m |
| (parameter setting) | 120000 $m^2$ | 20 | 10 | | Carrier frequency | 5.89 GHz |
| | | | | | Channel bandwidth | 6 Mbps |
| | | 20 | 10 | | PHY/MAC layer | IEEE 802.11p |
| U2 | 240000 $m^2$ | 30 | 15 | | Routing layer | OLSR |
| | | 40 | 20 | | Transport layer | UDP |
| | | | | | CBR packet size | 512 bytes |
| | | 30 | 15 | | CBR data rate | 33/66/100 333/666/1000 kbps |
| U3 | 360000 $m^2$ | 45 | 23 | | | |
| | | 60 | 30 | | CBR time | 60 s |

## 5.3 GA parameter setting experiments

A parameter setting analysis was performed to study the best values for the crossover probability ($p_C$) and the mutation probability ($p_M$) in the parallel GA. The analysis was done over a small VANET defined in scenario U1 (120000 m$^2$ and 20 vehicles, with reference values $E_{RFC} = 5680$ and $PDR_{RFC} = 88.23\%$). The population size of the parallel GA was fixed to 24 individuals, and the stopping criterion was set at 100 generations. The candidate values for the parameters were: $p_C$: 0.5, 0.7, 0.9; and $p_M$: 0.25, 0.0125, 0.006125.

Table 3 summarizes the parallel GA results for the nine combinations of $p_C$ and $p_M$ analyzed, reporting the average, relative standard deviation, and best values of fitness; the average energy and PDR, and the average gaps in energy and PDR with the standard RFC configuration (Equations 12 and 13). Fig. 5(a) presents the energy improvements with respect to the standard RFC configuration, and Fig. 5(b) compares the trade-offs between power consumption and PDR for each of the nine configurations studied.

$$GAP_{energy} = \frac{E_{RFC} - E(s)}{E_{RFC}} \quad (12) \qquad GAP_{PDR} = \frac{PDR_{RFC} - PDR(s)}{100} \quad (13)$$

Table 3  Experimental results: parameter setting for the parallel GA.

| ($p_C$, $p_M$) | fitness | | | metrics | | GAP | RFC |
|----------------|---------|-------|------|---------|------|-----|-----|
| | avg | stdev | best | energy | PDR | energy | PDR |
| (0.5,0.06125) | 0.576836 | 0.31% | 0.572319 | 3454.40 | 75.03% | 39.19% | -14.95% |
| (0.7,0.06125) | 0.577790 | 0.55% | 0.571034 | 3446.11 | 75.01% | 39.34% | -14.99% |
| (0.9,0.06125) | 0.577498 | 0.39% | 0.572754 | 3459.03 | 75.20% | 39.11% | -14.77% |
| (0.5,0.125) | 0.573733 | 0.21% | 0.571268 | 3447.76 | 75.03% | 39.31% | -14.95% |
| (0.7,0.125) | 0.573778 | 0.24% | 0.570946 | 3445.84 | 75.05% | 39.34% | -14.93% |
| (0.9,0.125) | 0.576217 | 0.14% | 0.574546 | 3470.34 | 75.33% | 38.91% | -14.61% |
| (0.5,0.25) | 0.574279 | 0.13% | 0.572724 | 3457.23 | 75.01% | 39.14% | -14.99% |
| (0.7,0.25) | 0.572346 | 0.15% | 0.570118 | 3440.33 | 75.01% | 39.44% | -14.99% |
| (0.9,0.25) | 0.572408 | 0.17% | 0.570351 | 3442.20 | 75.07% | 39.41% | -14.91% |

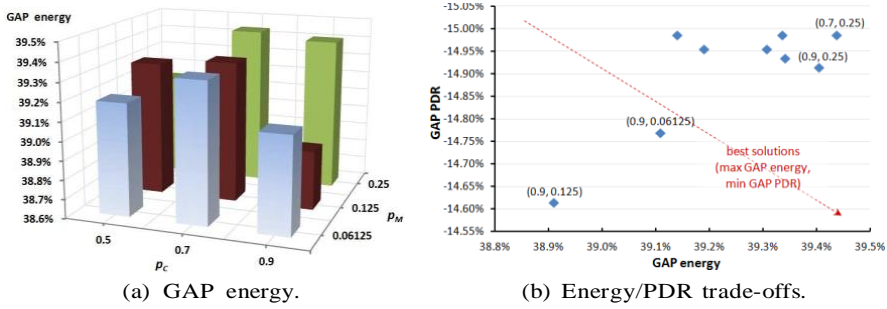

(a) GAP energy.                    (b) Energy/PDR trade-offs.

Fig. 5  Graphical summary: parameters setting for the parallel GA.

The graphic in Fig. 5(b) shows that four of the studied combinations of $p_C$ and $p_M$ obtained the best trade-off values between power consumption and PDR: (0.7, 0.25), (0.9, 0.25), (0.9, 0.06125), and (0.9, 0.125). Since this work is mainly concerned with reducing the power consumption, the most promising OLSR configurations are those in the far right section of the graphic in Fig. 5(b). When compared with the power consumption and PDR results obtained with the standard RFC configuration, the best results were obtained with the parameter configurations $p_C$=0.7, $p_M$=0.25.

### 5.4 Results and discussion

The experimental evaluation studied the quality of results and the computational efficiency of the parallel GA using the most promising parameter values identified in the previous subsection, to find an energy-aware configuration for OLSR in VANETs. In all the experiments reported in this subsection, the stopping criterion for the parallel GA was set at 500 generations.

The experimental analysis was performed over a medium-sized VANET defined in the scenario U2 (area 240 000 m², and involving 20 vehicles). The reference values for energy and PDR for this scenario are $E_{RF\ C} = 9104.19$ and $PDR_{RF\ C} = 87.12\%$.

#### 5.4.1 *Experimental results*

Table 4 summarizes the results of the experimental analysis over the mediumsized U2 scenario. Three parallel GA variants were studied: implementations using 8, 16, and 24 individuals, and the same number of execution threads. In order to provide a baseline for the comparison, the analysis includes the results obtained with two sequential optimization methods: a classic GA, using a population of 8 individuals and a single thread for execution, and the previous QoS optimized version of OLSR by means of Differential Evolution (DE-OLSR) [40].

Table 4 reports the average, relative standard deviation, and best fitness results obtained in 30 independent executions performed for each algorithm: parallel GA with 8 threads (pGA-8), parallel GA with 16 threads (pGA-16), and parallel GA with 24 threads (pGA-24). In addition, the power consumption and PDR values obtained with the best OLSR configuration found, and the gaps with respect to the standard RFC parametrization are also presented.

Table 4  Experimental results: parallel GA evaluation.

| algorithm | fitness | | | metrics | | GAP | RFC |
|---|---|---|---|---|---|---|---|
| | avg | stdev | best | energy | PDR | energy | PDR |
| sequential GA | 0.7521 | 2.66% | 0.7025 | 6909.12 | 80.48% | 24.11% | -6.64% |
| QoS DE-OLSR [40] | n/a | n/a | 0.7734 | 7798.48 | 97.55% | 14.34% | 10.43% |
| pGA-8 | 0.7058 | 1.88% | 0.6730 | 6551.89 | 74.74% | 28.03% | -12.38% |
| pGA-16 | 0.6883 | 1.69% | 0.6621 | 6446.80 | 75.20% | 29.19% | -11.92% |
| pGA-24 | 0.6774 | 1.37% | 0.6482 | 6305.58 | 75.14% | 30.74% | -11.98% |

In order to determine the significance of the comparison, a statistical analysis was performed over the results distributions for each parallel GA. First, the Kolmogorov-Smirnov test was applied to check whether the obtained fitness values follow a normal distribution or not. The $D$ metric values presented in the first row of Table 5 indicates that the results for pGA-8, pGA-16, and pGA-24 are not normally distributed. As a consequence, the non-parametric Kruskal-Wallis statistical test was performed with a confidence level of 95%, to compare the distributions for pGA-8, pGA-16, and pGA-24. The small $p$ values reported ($< 0.05$ in all cases) indicate that the fitness improvements can be considered statistically significant, thus the parallel GA using 24 threads is the best algorithm from among the studied methods.

Table 5  Statistical analysis of parallel GA results.

| statistical test | | algorithm | | |
|---|---|---|---|---|
| | | pGA-8 | pGA-16 | pGA-24 |
| Kolmogorov-Smirnov | | $< 10^{-7}$ | $< 10^{-7}$ | $< 10^{-7}$ |
| Kruskal-Wallis | pGA-8 | 6 | $6.4 \times 10^{-4}$ | $1.9 \times 10^{-7}$ |
| | pGA-16 | $.4 \times 10^{-4}$ | - | 0.015 |
| | pGA-24 | $1.9 \times 10^{-7}$ | 0.015 | - |

Overall, the results in Tables 4 and 5 demonstrate that significant improvements in the fitness values are computed using the parallel master-slave GA with 24 threads, when compared with the reference results from the sequential GA and DE-OLSR. The improvements in the fitness values bring forth   a significant decrease in the power consumption of the OLSR protocol: more than 30% of reduction with respect to the standard OLSR configuration was achieved for the best configuration found using pGA-24, while the PDR degradation remained below 12%.

The best energy-aware OLSR configuration—found by the parallel GA using 24 threads—is HELLO INTERVAL = 14.890, REFRESH INTERVAL = 7.416, TC INTERVAL = 28.158, WILLINGNESS = 5, NEIGHB HOLD TIME = 20.825, MID HOLD TIME = 10.814, TOP HOLD TIME = 70.959, and DUP HOLD TIME = 90.000.

The main advantages of this configuration are: i) it generates lower traffic control than the standard RFC configuration, since it increases the timeouts that control the protocol messages forwarding; ii) the power consumption of each vehicular node significantly decreases with respect to the one required when using the standard RFC configuration, because each node spends less time in the most consuming states (transmitting and receiving); and iii) all nodes show a higher will to act as MPR. On the other hand, a disadvantage of the proposed configuration is that it uses higher validity times, and therefore, it needs longer to detect link loss failures.

### 5.4.2 *Computational efficiency*

The most common metrics used by the research community to evaluate the performance of parallel algorithms are the *speedup* and the *efficiency*.

The speedup evaluates how much faster a parallel algorithm is than its corresponding sequential version. It is computed as the ratio of the execution times of the sequential algorithm ($T_1$) and the parallel version executed on $m$ computing elements ($T_m$) (Equation 14). When applied to non-deterministic algorithms, such as the parallel GA applied in this work, the speedup should compare the *mean* values of the sequential and parallel execution times (Equation 15) [4]. The ideal case for a parallel algorithm is to achieve linear speedup ($S_m = m$), but the most common situation is to achieve sublinear speedup ($S_m < m$), mainly due to the times required to communicate and synchronize the parallel processes.

The efficiency is the normalized value of the speedup, regarding the number of computing elements used to execute a parallel algorithm (Equation 16). This metric allows the comparison of algorithms eventually executed in nonidentical computing platforms. The linear speedup corresponds to $e_m = 1$, and in the most usual situations $e_m < 1$.

$$S_m = \frac{T_1}{T_m} \quad (14) \qquad S_m = \frac{E[T_1]}{E[T_m]} \quad (15) \qquad e_m = \frac{S_m}{m} \quad (16)$$

Table 6 compares the performance of the studied parallel GAs, showing the average and best execution times, and the values of the speedup and efficiency metrics when using 8, 16, and 24 threads. The results in Table 6 demonstrate that significants reductions in the required execution times are obtained when using the parallel GA implementations with respect to a sequential GA. Fig. 6 graphically summarizes the speedup and efficiency comparison for the three parallel GAs.

Table 6 Performance comparison of the proposed parallel GAs.

| algorithm | execution time (s) | | speedup | | efficiency | |
|---|---|---|---|---|---|---|
| | avg | best | avg | best | avg | best |
| parallel GA, 8 threads | 11113.73 | 9235.71 | 5.80 | 6.86 | 0.72 | 0.86 |
| parallel GA, 16 threads | 13192.70 | 12440.05 | 11.81 | 12.63 | 0.74 | 0.79 |
| parallel GA, 24 threads | 20239.02 | 13670.90 | 19.10 | 20.12 | 0.80 | 0.84 |

According to Amdahl's law [5], the performance of any parallel application is theoretically limited by the sequential part of the code, which mainly depends on the choice of the parallelization strategy. In the proposed parallel GAs, the fitness function evaluation is the most consuming part within the algorithm, since the VANET simulations using ns-2 demand large execution times. The results in Table 6 and Fig. 6 demonstrate that the proposed master-slave model is a useful choice to significantly reduce the execution times of the parallel GAs. Despite following a synchronous paradigm (that tends to generate idle times due to the synchronization of the execution threads), the parallel GAs show an almost-linear speedup behavior. The average efficiency values obtained were greater than 70% for the three implementations studied, and a maximum average of 80% was achieved when using the parallel GA with 24 threads.

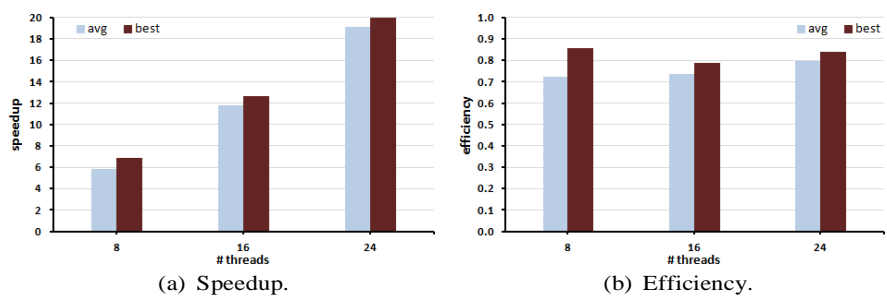

(a) Speedup.                    (b) Efficiency.

Fig. 6 Speedup and efficiency comparison for the parallel GAs.

## 5.5 Validation in other VANET scenarios

In order to confirm the efficacy of the results obtained in the experimental analysis, a set of validation experiments were conducted to compare the performance of the best OLSR configurations found using each parallel GA with the standard RFC configuration. The validation experiments involved simulations performed over 36 different unseen VANET scenarios, defined in the medium-size (U2) and large-size (U3) urban areas of Málaga, already presented in Section 5.1.

The validation analysis evaluated several metrics related to the energyaware and QoS of the communication. From the point of view of the power consumption, the energy in transmitting ($E_{send}$) and receiving ($E_{recv}$) mode, as well as the total energy ($E_{total}$) and total energy per vehicle ($E_{tot×v}$) were studied. From the point of view of QoS, the studied metrics include the PDR, the time spent until reaching the destination node (End-to-End Delay, E2ED, in miliseconds), the overload generated by the routing protocol (Normalized Routing Load, NRL), and the quality of the generated routing paths, evaluated by the number of hops required to reach the destination.

Table 7 presents for each best OLSR configuration found using the three parallel GAs studied, the average values for each studied metric, computed in the simulations performed over the 36 VANET scenarios. The results are compared with the reference values obtained in simulations performed with the standard OLSR configuration suggested by RFC 3626. The best average values obtained for each metric are marked in bold.

Table 7  Results of the validation experiments.

| config. | energy metrics | | | | QoS metrics | | | |
|---|---|---|---|---|---|---|---|---|
| | $E_{sent}$ | $E_{recv}$ | $E_{total}$ | $E_{tot×v}$ | PDR | E2ED | NRL | hops |
| *medium size (U2)* | | | | | | | | |
| pGA-8 | 12099.05 | 5265.45 | 17364.49 | 604.12 | 61.54% | 62.39 | 3.36% | 1.58 |
| pGA-16 | 11902.02 | 5206.53 | 17108.55 | 589.17 | 63.64% | 58.35 | 3.53% | 1.43 |
| pGA-24 | 11776.50 | 5094.87 | 16871.36 | 575.86 | 61.80% | 55.04 | 3.34% | 1.47 |
| RFC | 17918.45 | 8102.75 | 26021.20 | 876.91 | 70.22% | 1356.18 | 25.46% | 1.25 |
| *large size (U3)* | | | | | | | | |
| pGA-8 | 14682.85 | 7030.52 | 21713.36 | 491.22 | 55.75% | 505.30 | 3.98% | 1.50 |
| pGA-16 | 14864.78 | 7120.72 | 21985.51 | 505.51 | 57.63% | 490.34 | 3.73% | 1.48 |
| pGA-24 | 14249.18 | 6762.22 | 21011.39 | 479.16 | 56.65% | 483.62 | 3.57% | 1.45 |
| RFC | 21574.81 | 16247.10 | 37821.93 | 877.75 | 64.00% | 868.57 | 28.34% | 1.15 |
| *overall* | | | | | | | | |
| pGA8 | 13390.95 | 6147.99 | 19538.93 | 547.67 | 58.64% | 283.85 | 3.67% | 1.54 |
| pGA-16 | 13383.40 | 6163.63 | 19547.03 | 547.34 | 60.64% | 274.34 | 3.63% | 1.46 |
| pGA-24 | 13012.84 | 5928.54 | 18941.37 | 527.51 | 59.22% | 269.33 | 3.45% | 1.46 |
| RFC | 19572.25 | 12102.03 | 31674.29 | 877.33 | 67.89% | 506.26 | 25.22% | 1.20 |

### Power consumption

The values of the power consumption metrics in Table 7 indicate that significant reductions are obtained when using the OLSR parameterizations computed by using the three parallel GA. The configuration found by the parallel GA using 24 threads is the most efficient parametrization for OLSR in VANETs, allowing a reduction of up to 40.2% in the power consumption. This behavior was consistently verified in both transmitting and receiving communication modes, and in the overall energy utilization per vehicle.

Figure 7 presents the energy reductions with respect to the standard RFC configuration, regarding the dimension of the simulated scenarios.

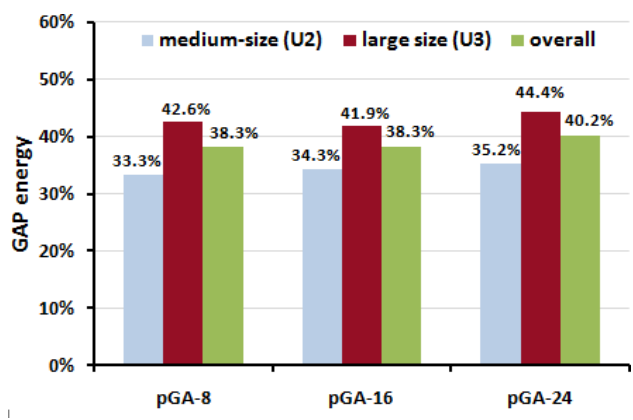

Fig. 7 Energy reductions with respect to the RFC, regarding the scenario dimension.

The results in Figure 7 demonstrate that significant improvements in the power consumption are obtained when using the configuration found with pGA-24. In addition, the energy reductions with respect to the standard RFC configuration increase for the largest scenarios simulated. The configuration found by pGA-24 achieved up to 44.4% of improvement in average for the larges scenarios, and a maximum value of 77.5% in a scenario with 40 vehicles. These notable improvements confirm previous claims about the inefficiency of the standard OLSR configuration in large VANET scenarios with high traffic density, already suggested by previous experimental evaluations [13].

The (non-parametric) Friedman statistical test was applied to analyze the comparison ranks between the energy results of pGA-8, pGA-16, pGA-24, and RFC. In addition, the Wilcoxon signed-rank statistical test was applied to analyze the mean ranks of the energy results, by evaluating the paired differences between the gaps values for all configurations. Table 8 summarizes the results of the statistical analysis. In the Wilcoxon test, the group of three values reported corresponds to the positive ranks, average positive ranks, and the sum of positive ranks for every pairwise comparison, respectively.

Table 8 Statistical analysis of the energy results.

| statistical test | | configuration | | | |
|---|---|---|---|---|---|
| | | pGA-8 | pGA-16 | pGA-24 | RFC |
| Friedman (*avg. rank*) | | 2.19 | 1.94 | 1.92 | 3.94 |
| Wilcoxon | pGA-8 | - | (14, 19.8, 277) | (16, 16.6, 266) | (35, 19.0, 665) |
| | pGA-16 | (22, 17.7, 389) | - | (16, 17.1, 274) | (36, 18.5, 666) |
| | pGA-24 | (20, 20.0, 400) | (20, 19.6, 392) | - | (35, 18.9, 661) |
| | RFC | (19.0, 1.0, 1) | (18.5, 0.0, 0) | (1, 5.0, 5) | - |

All the previous results demonstrate the efficacy of the proposed automatic methodology to compute accurate energy-aware OLSR configurations.

<u>Quality of service</u>

Regarding the QoS metrics, the results in Table 7 indicate that, when using the OLSR configuration computed by the parallel GA using 24 threads, the improvements in the power consumption are obtained without suffering large reductions in the PDR values—8% in average—. This is an acceptable value for the loss in the QoS, when taking into account the important energy reductions achieved.

An extremely large decrease is obtained in the transmission times required to reach the destination nodes (E2ED) when using the energy-aware OLSR configuration. This result is mainly motivated by the absence of congestion, due to the low overload generated. The NRL values indicate that all configurations found using the parallel GA exchange significantly less control messages than the standard OLSR. In average, the network overload is 1/7 of the standard one, showing that OLSR employing the automatic configuration is less likely to be affected by network congestion problems than the standard OLSR. This feature allows the new configuration to be more useful than the standard one in situations where a large number of messages are transmitted, such as in city center areas, traffic jam scenarios, etc. However, the values of the hops metric indicates that the standard OLSR finds shorter paths than the energyaware OLSR. Anyway, the routing paths computed by the energy-aware OLSR do not use longer than 1.5 hops in average to reach the destination node, while the RFC configuration requires 1.20.

The previously commented QoS results indicate that the automatic energyaware OLSR configuration found by pGA-24, while keeping the PDR degradations under a controlled threshold, generates less network routing overload, and it also allows a faster delivery of the packets. The standard OLSR computes shorter routing paths, but the size difference with the routing paths computed with the new energy-aware OLSR is negligible, so both configurations can be considered as equivalent regarding this metric. Indeed, the standard configuration is much more congestion-prone due to the large network overload and collisions.

<u>Experimental analysis: summary</u>

The experimental analysis proved that the energy-aware OLSR configuration is able to obtain large reductions in the power consumption and significantly improve the time required to deliver the data packets, while only suffering a bounded degradation in the PDR metric. The relevance of all the considerations commented on the previous subsections increase when facing large-sized VANET scenarios where real-time transmissions are important, such as in traffic accidents, traffic jams, urban areas with high density of VANET users, etc. In these situations, the results obtained demonstrate the efficacy of the proposed automatic method for finding energy-aware OLSR configurations.

## 6 Conclusions

This article has studied the problem of finding energy-efficient configurations for the OLSR routing protocol in vehicular networks. The design of energy-efficient communication protocols is an important issue in this research area, and few previous researches have tackled the OLSR configuration problem from an energy-oriented point of view. In this line of research, the main contribution of this article is to propose an automatic methodology for computing energy-efficient configurations for the OLSR protocol in VANETs, by using a parallel GA.

The automatic search for energy-aware OLSR configurations is carried out by considering the power consumption of the VANET nodes as the main objective to optimize, but also taking into account the level of QoS in the communications. A well-know energy model in wireless networks and the ns-2 network simulator were used. The proposed GA for solving the problem applies a master-slave parallel model. It enables the configurations search to be performed efficiently, by simultaneously using several computing resources to perform the VANET simulations. By reducing the execution times, the parallel GA allows increasing the population of candidate solutions in order to overcome the stagnation problem identified in previous proposals. The computational efficiency of the proposed parallel GA was almost-linear, obtaining efficiency values greater than 80%.

Regarding the wireless communications, the experimental analysis demonstrates that significant reductions in the power consumption of the VANET nodes are obtained when using the automatic energy-aware OLSR configuration found by the parallel GA, when compared with the standard OLSR configuration suggested by RFC 3626. Average reductions up to 40.2% in the power consumption were obtained, and significantly better improvements (up to 77.54%) were computed for large and dense VANET scenarios. In addition, the energy-aware OLSR configuration found significantly reduces the network overload, and thus it allows reducing the average time required to deliver the data packets. All these important features are obtained while only suffering a bounded degradation (less than 8%) in the QoS of the communication, evaluated by the PDR metric.

The main lines for future work are related to two issues: improving the method used in the automatic search, and tackling the OLRS configuration as a multiobjective problem. Regarding the first issue, the use of new fitness functions should be considered, taking into account new power-aware and QoS metrics, such as the residual level of battery of the nodes and the packet delays, respectively. In addition, the approach proposed in this paper could be extended by using several VANET scenarios to evaluate each OLSR configuration, possibly by using other efficient models for parallel EAs. Thus, different situations will be taken into account to obtain more accurate fitness results. Regarding the second issue, the study of explicit multiobjective approaches for the problem is also suggested as future work, in view that the OLSR energy savings vary in inverse proportion with the QoS of the protocol.

Acknowledgements J. Toutouh is supported by grant AP2010-3108 from the Spanish Government. The work of S. Nesmachnow has been partially supported by ANII and PEDECIBA, Uruguay. The work of J. Toutouh and E. Alba has been partially funded by the Spanish Ministry MICINN and FEDER under contracts TIN2008-06491-C04-01 (M* project) and TIN2011-28194 (roadME project), and CICE, Junta de Andaluc´ıa, under contract P07-TIC-03044 (DIRICOM project).

## References

1. Alba, E.: Parallel Metaheuristics: A New Class of Algorithms. Wiley-Interscience (2005)
2. Alba, E., Almeida, F., Blesa, M., Cotta, C., Diaz, M., Dorta, I., Gabarr´o, J., Gonz´alez, J., Le´on, C., Moreno, L., Petit, J., Roda, J., Rojas, A., Xhafa, F.: MALLBA: A library of skeletons for combinatorial optimisation. Parallel Computing 32(5-6), 415–440 (2006)
3. Alba, E., Dorronsoro, B., Luna, F., Nebro, A., Bouvry, P., Hogie, L.: A Cellular MOGA for Optimal Broadcasting Strategy in Metropolitan MANETs. Computer Communications 30(4), 685–697 (2007). DOI 10.1109/IPDPS.2005.4
4. Alba, E., Tomassini, M.: Parallelism and evolutionary algorithms. IEEE Trans. Evol. Comput. 6(5), 443–462 (2002)
5. Amdahl, G.: Validity of the single processor approach to achieving large scale computing capabilities. In: Proceedings of the Spring Joint Computer Conference, AFIPS '67, pp. 483–485. ACM (1967)
6. B¨ack, T., Fogel, D., Michalewicz, Z. (eds.): Handbook of evolutionary computation. Oxford University Press (1997)
7. Benslimane, A., El Khoury, R., El Azouzi, R., Pierre, S.: Energy power-aware routing in OLSR protocol. In: Proceedings of the 1st Mobile Computing and Wireless Communication International Conference, pp. 14–19 (2006)
8. Cano, J., Manzoni, P.: A performance comparison of energy consumption for mobile ad hoc network routing protocols. In: Proceedings of the 8th International Symposium on Modeling, Analysis and Simulation of Computer and Telecommunication Systems, pp. 57–64. IEEE Computer Society (2000)
9. Chen, T., Mehani, O., Boreli, R.: Trusted routing for VANET. In: M. Berbineau, M. Itami, G. Wen (eds.) ITST 2009, 9th International Conference on Intelligent Transport Systems Telecommunications, pp. 647–652. IEEE Computer Society, Piscataway, NJ, USA (2009)
10. Cheng, H., Yang, S.: Genetic algorithms with immigrant schemes for dynamic multicast problems in mobile ad hoc networks. Eng. Appl. Artif. Intell. 23, 806–819 (2010)
11. Chou, C., Chen, J.: Genetic algorithms: initialization schemes and genes extraction. In: The Ninth IEEE International Conference on Fuzzy Systems, vol. 2, pp. 965–968 (2000)
12. Clausen, T., Jacquet, P.: Optimized Link State Routing Protocol. IETF RFC 3626, [online] Available in http://www.ietf.org/rfc/rfc3626.txt (2003). Retrieved October 2011
13. De Rango, F., Cano, J., Fotino, M., Calafate, C., Manzoni, P., Marano, S.: OLSR vs DSR: A comparative analysis of proactive and reactive mechanisms from an energetic point of view in wireless ad hoc networks. Computer Communications 31(16), 3843–3854 (2008)
14. De Rango, F., Fotino, M.: Energy efficient OLSR performance evaluation under energy aware metrics. In: Proceedings of the 12th international conference on Symposium on Performance Evaluation of Computer & Telecommunication Systems, SPECTS'09, pp. 193–198. IEEE Press, Piscataway, NJ, USA (2009)
15. Dorronsoro, B., Danoy, G., Bouvry, P., Alba, E.: Evaluation of different optimization techniques in the design of ad hoc injection networks. In: Workshop on Optimization Issues in Grid and Parallel Computing Environments, part of the HPCS, pp. 290–296. Nicossia, Cyprus (2008)
16. Feeney, L.M., Nilsson, M.: Investigating the energy consumption of a wireless network interface in an ad hoc networking environment. In: In IEEE Infocom, pp. 1548–1557 (2001)

17. Garc´ıa-Nieto, J., Alba, E.: Automatic parameter tuning with metaheuristics of the AODV routing protocol for vehicular ad-hoc networks. In: C.D. Chio, A. Brabazon, G.A.D. Caro, M. Ebner, M. Farooq, A. Fink, J. Grahl, G. Greenfield, P. Machado, M. O'Neill, E. Tarantino, N. Urquhart (eds.) EvoApplications (2), *Lecture Notes in Computer Science*, vol. 6025, pp. 21–30. Springer (2010)
18. Garc´ıa-Nieto, J., Toutouh, J., Alba, E.: Automatic tuning of communication protocols for vehicular ad hoc networks using metaheuristics. Engineering Applications of Artificial Intelligence 23(5), 795–805 (2010)
19. Ge, Y., Kunz, T., Lamont, L.: Quality of service routing in ad-hoc networks using OLSR. In: Proceedings of the 36th Annual Hawaii International Conference on System Sciences, p. 300. IEEE Computer Society (2003). [electronic publication]
20. Ghanem, N., Boumerdassi, S., Renault, E.: New energy saving mechanisms for mobile ad-hoc networks using OLSR. In: Proceedings of the $2^{nd}$ ACM International Workshop on Performance Evaluation of Wireless Ad Hoc, Sensor, and Ubiquitous Networks, pp. 273–274. ACM (2005)
21. Goldberg, D.E.: Genetic Algorithms in Search Optimization and Machine Learning. Addison-Wesley (1989)
22. Guo, Z., Malakooti, B.: Energy aware proactive MANET routing with prediction on energy consumption. In: Proceedings of the International Conference on Wireless Algorithms, Systems and Applications, pp. 287–293. IEEE Computer Society (2007)
23. H¨arri, J., Filali, F., Bonnet, C.: Performance comparison of AODV and OLSR in VANETs urban environments under realistic mobility patterns. In: Med-Hoc-Net 2006, 5th Annual Mediterranean Ad Hoc Networking Workshop. IFIP (2006)
24. Hartenstein, H., Laberteaux, K.: VANET Vehicular Applications and Inter-Networking Technologies. Intelligent Transport Systems. John Wiley & Sons, Upper Saddle River, NJ, USA (2009)
25. Huhtonen, A.: Comparing AODV and OLSR routing protocols. In: Telecommunications Software and Multimedia, pp. 1–9 (2004)
26. Krajzewicz, D., Bonert, M., Wagner, P.: The open source traffic simulation package SUMO. In: RoboCup'06, pp. 1–10 (2006)
27. Kunz, T.: Energy-efficient MANET routing: Ideal vs. realistic performance. In: International Wireless Communications and Mobile Computing Conference, pp. 786 – 793 (2008)
28. Laouiti, A., Mu¨hlethaler, P., Sayah, F., Toor, Y.: Quantitative evaluation of the cost of routing protocol OLSR in a Vehicle Ad Hoc NETwork (VANET). In: VTC Spring, pp. 2986–2990. IEEE (2008)
29. Lee, K.C., Lee, U., Gerla, M.: Survey of Routing Protocols in Vehicular Ad Hoc Networks, chap. 8, pp. 149–170. Eds. IGI Global (2009)
30. Li, F., Wang, Y.: Routing in vehicular ad hoc networks: A survey. IEEE Vehicular Technology Magazine 2(2), 12–22 (2007)
31. Mahfoudh, S., Minet, P.: An energy efficient routing based on OLSR in wireless ad hoc and sensor networks. In: Proceedings $22^{nd}$ International Conference on Advanced Information Networking and Applications, pp. 1253–1259. IEEE Computer Society (2008)
32. Nguyen, D., Minet, P.: Analysis of MPR selection in the OLSR protocol. Advanced Information Networking and Applications Workshops, International Conference on 2, 887–892 (2007)
33. Razalli, S., Wong, K., Suhaimi, S.: Enhancing the Willingness on the OLSR Protocol to Optimize the Usage of Power Battery Power Sources Left. International Journal of Engineering 2, 12–26 (2008)
34. Ros, F.J.: UM-OLSR: OLSR implementation for ns2. [online] Available in http://masimum.dif.um.es/?Software:UM-OLSR. Retrieved October 2011
35. Ruiz, P., Dorronsoro, B., Bouvry, P.: Optimization and performance analysis of the AEDB broadcasting algorithm. In: Computer Communications and Networks (ICCCN), 2011 Proceedings of 20th International Conference on, pp. 1 –6 (2011)
36. Ruiz, P., Dorronsoro, B., Valentini, G., Pinel, F., Bouvry, P.: Optimisation of the enhanced distance based broadcasting protocol for manets. The Journal of Supercomputing pp. 1–28 (2011)
37. Sangeeta, K., Sing, K.: Energy Efficient Routing In MANET Using OLSR. International Journal on Computer Science and Engineering 3(16), 1418–1421 (2011)

38. Santa, J., Tsukada, M., Ernst, T., Mehani, O., G´omez-Skarmeta, A.F.: Assessment of VANET multi-hop routing over an experimental platform. Int. J. Internet Protoc. Technol. 4(3), 158–172 (2009)

39. Spaho, E., Barolli, L., Mino, G., Xhafa, F., Kolici, V., Miho, R.: Performance evaluation of AODV, OLSR and DYMO protocols for vehicular networks using CAVENET. In: Network-Based Information Systems (NBiS), 2010 13th International Conference on,
pp. 527 –534 (2010). DOI 10.1109/NBiS.2010.79

40. Toutouh, J., Alba, E.: An efficient routing protocol for green communications in vehicular ad-hoc networks. In: Proceedings of 13th Annual Genetic and Evolutionary Computation Conference, GECCO 2011, pp. 719–726. ACM (2011)

41. Toutouh, J., Alba, E.: Optimizing OLSR in VANETs with Differential Evolution: A Comprehensive Study. In: First ACM International Symposium on Design and Analysis of Intelligent Vehicular Networks and Applications (DIVANet '11), DIVANet. ACM (2011)

42. Toutouh, J., Garc´ıa-Nieto, J., Alba, E.: Optimal configuration of OLSR routing protocol for VANETs by means of Differential Evolution. In: 3$^{rd}$ International Conference on Metaheuristics and Nature Inspired Computing, p. 8 (2010)

43. Unex: DCMA-86P2 Network Interface Card. [online] Available in http://www.unex.com.tw/product/dcma-86p2. Retrieved January 2012